\documentclass{article}
\usepackage{amssymb}
\usepackage{graphicx}
\usepackage{booktabs}
\usepackage{multirow}
\usepackage{array}
\usepackage{wrapfig}
\usepackage{enumitem}
\usepackage{float}
\usepackage{caption}
\usepackage[table]{xcolor}
\usepackage[preprint]{corl_2025} % Uncomment for the camera-ready ``final'' version.
\usepackage{algorithm}
\usepackage{algpseudocode}
\usepackage{amsmath}

\usepackage[most]{tcolorbox}
\usepackage{bm}
\usepackage{amsmath}
\usepackage{amsthm}
\usepackage{amsfonts}
\usepackage{wasysym}
\newcommand{\Xmark}{$\times$}%

\DeclareMathAlphabet{\mathpzc}{OT1}{pzc}{m}{it}

\DeclareFontFamily{U}{jkpmia}{}
\DeclareFontShape{U}{jkpmia}{m}{it}{<->s*jkpmia}{}
\DeclareFontShape{U}{jkpmia}{bx}{it}{<->s*jkpbmia}{}
\DeclareMathAlphabet{\mathfrak}{U}{jkpmia}{m}{it}

\newcommand{\bF}{\bm{F}}

\newcommand{\bx}{\bm{x}}

\newcommand{\ba}{\bm{a}}

\newcommand{\bn}{\bm{n}}

\newcommand{\N}{\mathcal{N}}

\definecolor{black}{rgb}{0.3,0.3,0.3}
\definecolor{blackb}{rgb}{0.3,0.3,0.3}
\definecolor{brownb}{rgb}{0.3,0.3,0.3}

\newtcbtheorem[no counter]{Definition}{Definition}{
  enhanced,
  rounded corners,
  attach boxed title to top left={
    yshifttext=-1mm
  },
  colback=white,
  colframe=black,
  fonttitle=\bfseries,
  coltitle=white,
  boxrule=0.5pt,
  boxed title style={
    rounded corners,
    size=small,
    colback=black,
    colframe=black,
  } 
}{def}

\title{\Large{Multi-Loco: Unifying Multi-Embodiment Legged Locomotion via Reinforcement Learning Augmented Diffusion}}

\author{
  Shunpeng Yang$^{*1}$,
  Zhen Fu$^{*1}$,
  Zhefeng Cao$^{1}$,
  Junde Guo$^{1}$, \AND
  Patrick Wensing$^{2}$,
  Wei Zhang$^{1,4}$,
  Hua Chen$^{3,4}$\\
  $^1$ Southern University of Science and Technology, China. $^2$ University of Notre Dame, United States \\
  $^3$ Zhejiang University-University of Illinois Urbana-Champaign Institute, China \\
  $^4$LimX Dynamics, Shenzhen, China. $^*$ Equal contribution \\[0.3ex]
}

\begin{document}
\maketitle

%===============================================================================
\begin{figure*}[h]
    \vspace{-1.1cm}
    \centering
    \includegraphics[width=0.9\linewidth]{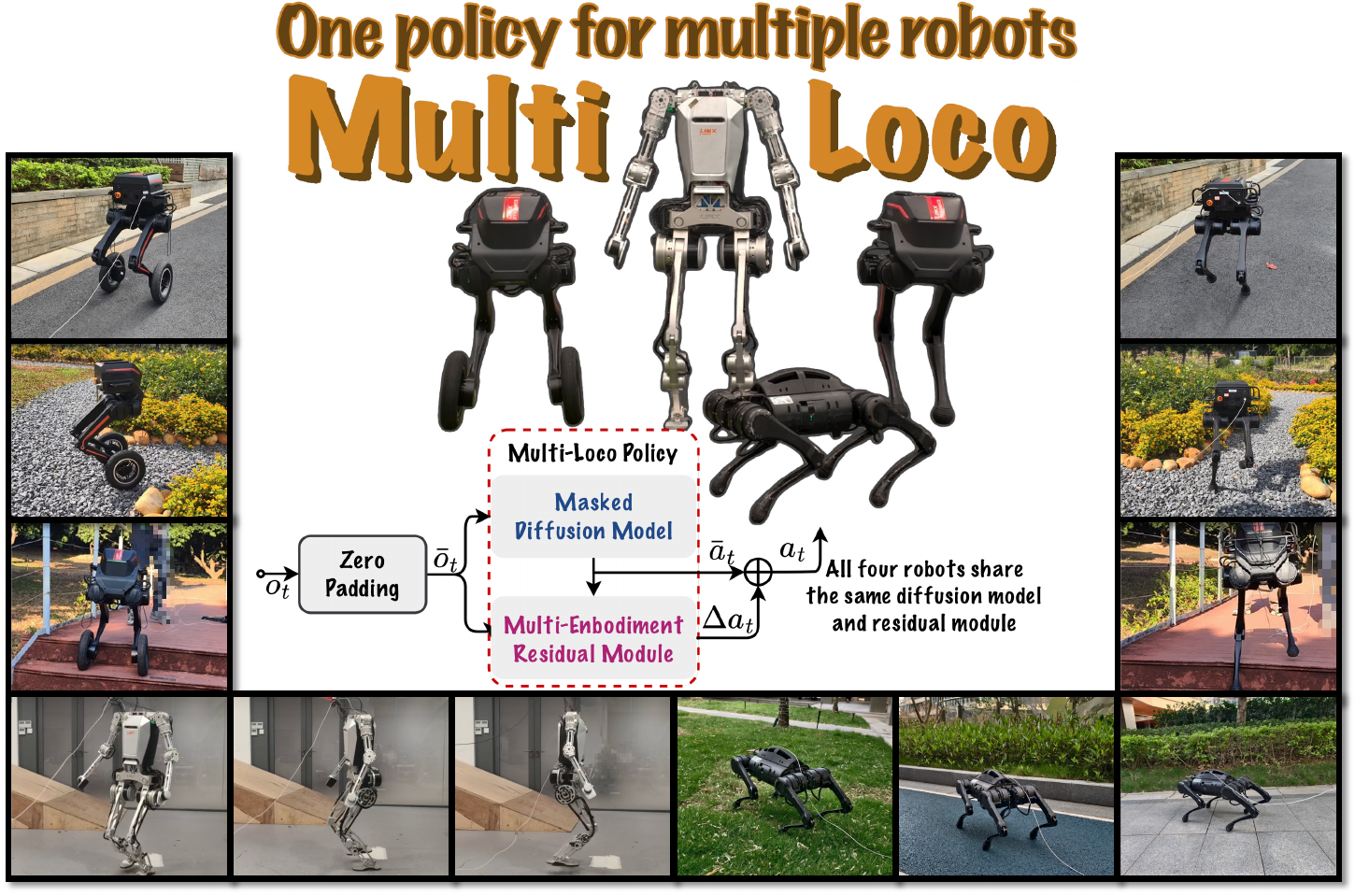}
    % \vspace{-0.3cm}
    \caption{\footnotesize{Deployment of the reinforcement learning augmented diffusion policy on four platforms (biped, wheeled biped, humanoid and quadruped). The experimental results demonstrate that the unified policy can effectively control the robots across various types of uneven terrain, including grass, slopes, stairs, and gravel paths. These results highlight the policy's robustness and exceptional control capabilities.}}
    \label{fig:real world experiments}
    \vspace{-0.5cm}
\end{figure*}
\begin{abstract}
    Generalizing locomotion policies across diverse legged robots with varying morphologies is a key challenge due to differences in observation/action dimensions and system dynamics. In this work, we propose \textit{Multi-Loco}, a novel unified framework combining a morphology-agnostic generative diffusion model with a lightweight residual policy optimized via reinforcement learning (RL). The diffusion model captures morphology-invariant locomotion patterns from diverse cross-embodiment datasets, improving generalization and robustness. The residual policy is shared across all embodiments and refines the actions generated by the diffusion model, enhancing task-aware performance and robustness for real-world deployment. We evaluated our method with a rich library of four legged robots in both simulation and real-world experiments. Compared to a standard RL framework with PPO, our approach - replacing the Gaussian policy with a diffusion model and residual term - achieves a 10.35\% average return improvement, with gains up to 13.57\% in wheeled-biped locomotion tasks. These results highlight the benefits of cross-embodiment data and composite generative architectures in learning robust, generalized locomotion skills. Project website: \url{https://multi-loco.github.io}
\end{abstract}
% Two or three meaningful keywords should be added here
%===============================================================================

\section{Introduction}
Generalizing locomotion policies across legged robots with diverse embodiments is a fundamental challenge in robotics, but it offers a promising path toward scalable, efficient learning by enabling knowledge reuse and reducing platform-specific engineering.
%Generalizing locomotion policies across legged robots with diverse embodiments remains a fundamental challenge in robotics, which is a key step toward scalable and efficient robot learning, enabling knowledge reuse across platforms and reducing robot-specific engineering effort. 
Although reinforcement learning (RL) has led to impressive progress in training locomotion controllers~\citep{gu2024advancing, hwangbo2019learning, lee2020learning, miki2022learning}, these policies are typically trained specifically for each robot morphology. Even robots with similar kinematic characteristics require separate training pipelines, limiting scalability and knowledge reuse. As a result, locomotion policies and datasets remain siloed, restricting unified learning across embodiments. While direct policy transfer to unseen robots remains an open problem, enabling policy generalization across diverse robots is an essential first step toward scalable multi-robot learning.

Recently, cross-embodiment learning has been attracting considerable research attention~\citep{yang2023polybot, xu2023xskill, team2024octo, doshi2024scaling}. These pioneering works have been conducted primarily in the domain of robotic manipulation tasks, aiming to develop generalizable policies that can handle variations in kinematics, sensing modalities, and control interfaces. However, transferring this success to legged locomotion presents unique challenges. Unlike manipulation, locomotion is deeply influenced by dynamical characteristics of the robots as well as their physical interaction with the environment, making it more difficult to abstract away embodiment-specific features. Moreover, reconciling the variations in observation and action dimensions across embodiments adds further complexity to the model design.

In this work, we propose \textit{Multi-Loco}, a novel unified framework that addresses multi-embodiment locomotion learning through the integration of a generative diffusion model and a residual RL policy. Rather than relying on end-to-end transformer-based regression methods, we use a diffusion model to learn a morphology-invariant policy from diverse locomotion datasets, capturing generalizable patterns across robot embodiments. The output of the diffusion model is then refined by a lightweight residual policy trained with reinforcement learning, enhancing task-specific performance and adaptation, while remaining shared across all embodiments.

In summary, our key contributions are as follows:
\begin{itemize}[leftmargin=12pt]
\item We propose a novel framework based on the integration of generative models with reinforcement learning to accommodate the design of unified locomotion policies across diverse legged platforms. Our approach captures shared locomotion patterns across various robots via leveraging the multi-modal capabilities of diffusion, and further bridges the sim2real gap with a shared residual policy trained via reinforcement learning.
\item We leverage the multi-modal nature of diffusion models to conditionally model locomotion behaviors across diverse embodiments. By aligning heterogeneous observation and action spaces through zero padding and employing masked score matching during training, our framework enables a single generative policy to generalize across multiple robot platforms.
\item We introduce a shared residual reinforcement learning policy that refines the diffusion model’s action outputs to further bridge the sim2real gap. By specifying task-aware rewards and employing a multi-critic architecture, the single residual policy is capable of enhancing locomotion performance across the given embodiments, complementing the generative prior without requiring embodiment-specific tuning.
\end{itemize}

\section*{Related Work}
\vspace{-0.3cm}
{\bf Reinforcement learning for legged locomotion:} 
RL has emerged as a powerful approach in the synthesis of robust and adaptable control policies in robotics, particularly for complex legged locomotion~\citep{gu2024advancing, hwangbo2019learning, lee2020learning, miki2022learning, wang2024cts, cheng2024extreme}. Hoeller et al.~\citep{hoeller2024anymal} demonstrated RL's capability to achieve agile parkour behaviors with quadrupedal robots, while Lee et al.~\citep{lee2024learning} extended RL to wheeled-legged systems, tackling the combined challenges of locomotion and navigation. RL has also been applied to whole-body humanoid control. OmniH2O~\citep{he2024omnih2o} combined teleoperation and RL to achieve dexterous full-body manipulation, and Hover~\citep{he2024hover} used neural controllers to unify locomotion and manipulation tasks. However, most of these methods require embodiment-specific training, with limited generalizability across different platforms. The integration of RL with advanced generative models, such as diffusion models, to enhance cross-embodiment versatility remains underexplored.

{\bf Diffusion models and their applications in robotics:} Diffusion models refine noisy samples through a learned denoising process, approximating complex, high-dimensional distributions. Song et al.~\citep{song2020score} established a theoretical foundation connecting diffusion to stochastic differential equations (SDEs), enabling accelerated sampling via ODE solvers~\citep{song2020denoising, lu2022dpm} and further inspiring frameworks such as the Elucidated Diffusion Model (EDM)~\citep{edm}. In robotics, diffusion models have shown promising applications for their ability to represent multi-modal behavior. Chi et al.~\citep{chi2023diffusion} applied them to contact-rich manipulation tasks, while Huang et al.~\citep{huang2024diffuseloco} introduced DiffuseLoco, one of the first applications of diffusion models to legged locomotion, demonstrating offline training and robust online control abilities. The potential of diffusion models for cross-embodiment locomotion learning remains to be further explored.

{\bf Cross-embodiment learning of legged robots:} 
Cross-embodiment learning aims to develop control policies that generalize across robots with varying morphologies, actuation, and sensing. Early work in this field focused on morphology-specific adaptation, such as GenLoco~\citep{feng2023genloco}, which trained quadruped controllers through procedural morphology randomization. However, such an approach was limited to robots with fixed degrees of freedom (DoFs). Subsequent investigations such as ManyQuadrupeds~\citep{shafiee2024manyquadrupeds} generalizes to diverse quadruped morphologies by combining Central Pattern Generators (CPGs) with reinforcement learning, though their task-space control paradigm requires predefined robot-specific inverse kinematics. To address morphological variability, MorAL~\citep{luo2024moral} introduced a morphology-aware network (MorphNet) that encodes physical information from proprioceptive observations. By conditioning policy learning on a compact morphology representation, MorAL improved robustness and generalization. Such an approach still relied on explicit morphology embeddings. Recently, URMA~\citep{bohlinger2024one} employed transformers to unify control across quadrupeds, bipeds, and hexapods via morphology-agnostic encoders, demonstrating effective transfer with explicit joint descriptions for decoding. CrossFormer~\citep{doshi2024scaling} further investigated generalist policies across legged platforms, yet their reliance on structured embodiment descriptors or per-robot modules may hinder scalability to new morphologies. 

\begin{table}[htbp!]
    \vspace{-0.5cm}
    \centering
    \caption{\footnotesize{Comparison of cross-embodiment learning models}}
    \begin{tabular}{lcccc}
    \hline
          Method &  \shortstack{Morphology \\ Agnostic} & \shortstack{Independent of \\ Joint Information} 
          & \shortstack{Cross \\ Tasks}  & \shortstack{Generative\\ Model} \\
    
    \hline
        GenLoco~\citep{feng2023genloco} & \Xmark & \checked  & \Xmark & \Xmark \\
        ManyQuadrupeds~\citep{shafiee2024manyquadrupeds} & \Xmark & \Xmark  & \Xmark & \Xmark \\
        MorAL~\citep{luo2024moral} & \Xmark & \Xmark  & \Xmark & \Xmark \\
        URMA~\citep{bohlinger2024one} & \checked & \Xmark  & \Xmark & \Xmark \\
        CrossFormer~\citep{doshi2024scaling} & \checked & \Xmark &  \checked & \Xmark \\
    \emph{Multi-Loco} (proposed) & \checked & \checked &  \Xmark & \checked \\
    \hline
    \end{tabular}
    \label{table: cross-embodiment models}
    \vspace{-0.3cm}
\end{table}

While these approaches have shown promising results in multi-embodiment learning, they often incorporate explicit morphology information - such as structural descriptors or per-robot observation encoders - to facilitate generalization. This design choice may present challenges for scaling to new robots or for learning in more morphology-agnostic settings. In this context, frameworks that aim to unify control across embodiments without relying on embodiment-specific inputs or architecture components remain relatively less explored.

\begin{figure}
    \centering
    \includegraphics[width=1.0\linewidth]{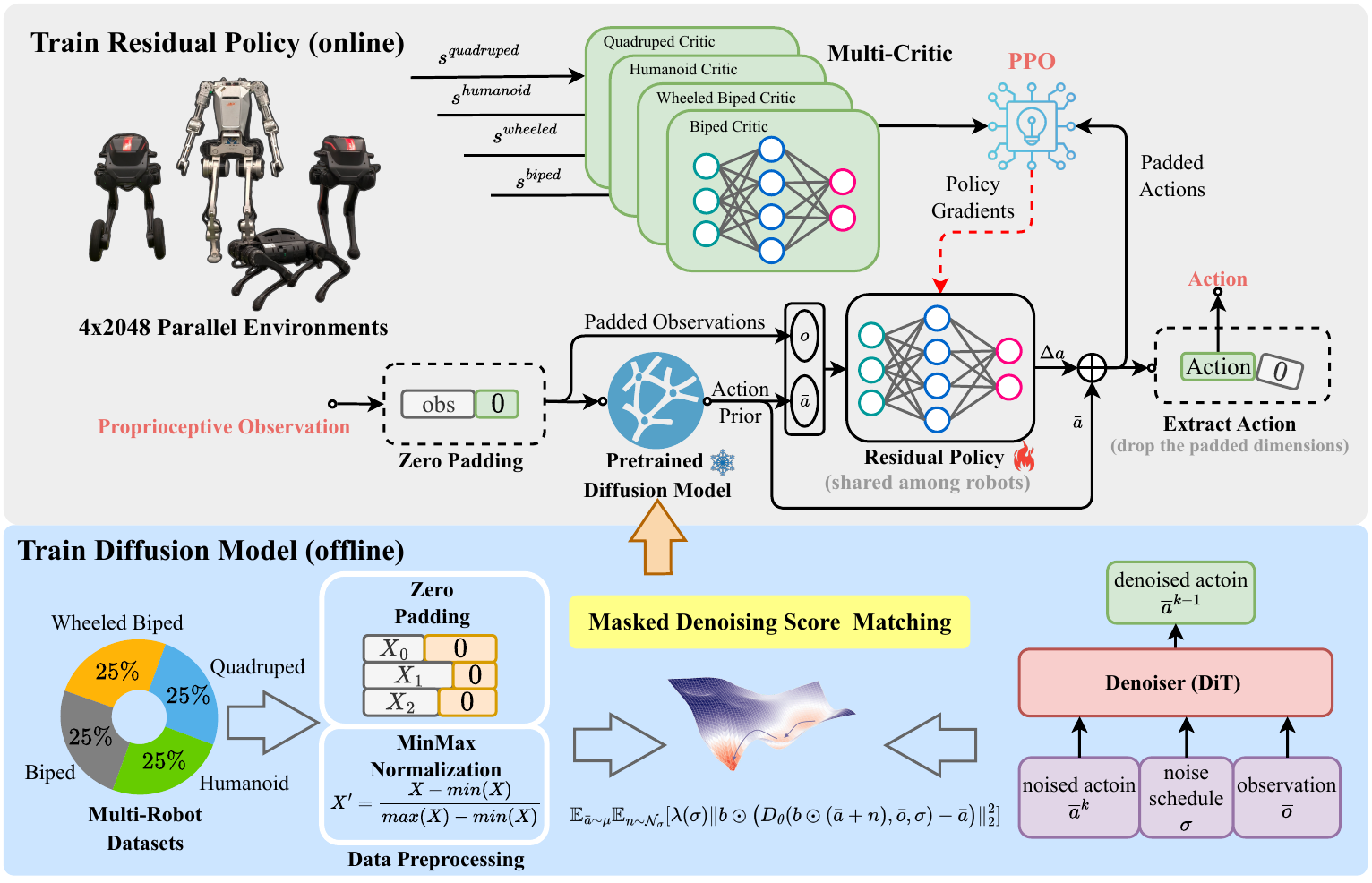}
    \caption{\footnotesize{
    Overview of the \textit{Multi-Loco} framework. Multi-robot datasets are preprocessed via zero-padding and normalization to align observation and action spaces across embodiments. A shared diffusion model is trained offline using masked denoising score matching. At inference time, the diffusion model generates action priors, which are refined by a residual policy trained via multi-critic PPO. Each critic specializes in one robot type, while the policy remains shared across all embodiments.
    }}
    \vspace{-0.5cm}
    \label{fig:framework}
\end{figure}

\section{Methodology}
We propose \textit{Multi-Loco}, a unified policy framework for multi-embodiment locomotion that combines generative modeling and reinforcement learning. Without relying on explicit morphology inputs, Multi-Loco captures shared locomotion patterns across diverse robots through a common policy structure. The framework consists of three components: (1) dimension alignment via zero-padding; (2) a diffusion model trained with masked denoising score matching; and (3) a residual reinforcement learning policy with multi-critic architecture. Details are provided below..

\subsection{Dimension Alignment for Multi-Embodiment Observations and Actions}

Multi-embodiment learning presents a fundamental challenge due to variations in observation and action spaces across different embodiments. To unify the policy space, we align all robot data into common observation and action spaces via zero-padding
\begin{equation*}
    \dim(\bar{\mathcal{O}}) = \max_{m \in \mathcal{M}} \dim(\mathcal{O}_m), \quad
    \dim(\bar{\mathcal{A}}) = \max_{m \in \mathcal{M}} \dim(\mathcal{A}_m)
\end{equation*}
where $\mathcal{M}$ denotes the set of robot embodiments, and $\bar{\mathcal{O}}$, $\bar{\mathcal{A}}$ are the unified observation and action spaces, respectively. For any robot $m$, its proprioceptive observation $\bm{o}_m \in \mathcal{O}_m$ and action $\bm{a}_m \in \mathcal{A}_m$ are mapped into $\bar{\bm{o}}_m \in \bar{\mathcal{O}}$ and $\bar{\bm{a}}_m \in \bar{\mathcal{A}}$ by padding zeros to the unused dimensions. In parallel, we define a binary mask $\bm{b} \in \{0,1\}^{\dim(\bar{\mathcal{A}})}$ that indicates which action dimensions are valid (1) or padded (0). For example, if a quadruped robot has a 12-dimensional action space and $\dim(\bar{\mathcal{A}})=20$, then its mask is $\bm{b} = [\underbrace{1,\dots,1}_{12},\underbrace{0,\dots,0}_{8}]$.

To improve training stability across heterogeneous embodiments, we further apply MinMax normalization to all observation and action dimensions. This normalization ensures consistent numerical ranges across robots, facilitating better convergence during diffusion model training and allowing shared model parameters to generalize effectively across varied scales.

This preprocessing standardizes observation-action formats while preserving the structural differences between embodiments. The resulting unified dataset forms the input for training a generative policy model, as described in the next section.

\subsection{Multi-Embodiment Locomotion Unification via Diffusion Model}

Learning robust policies across embodiments requires addressing multi-modality in robot dynamics and control strategies. To address this challenge, we propose employing diffusion models to directly parameterize the robot's action distribution $\mu(\bar{\bm{a}}|\bar{\bm{o}})$ , leveraging their demonstrated advantages in sample quality and robustness over alternative approaches such as VAEs and flow models, defined over the robot's action space $\bar{\bm{a}}\in\bar{\mathcal{A}}$ and proprioceptive observation space $\bar{\bm{o}}\in\bar{\mathcal{O}}$.

To ensure real-time inference, we adopt the EDM with a lightweight Diffusion Transformer (DiT)~\cite{dit} backbone. EDM approximates the Stein score function $\nabla_{\bar{\bm{a}}}\log \mu_\sigma(\bar{\bm{a}}|\bar{\bm{o}})$ and reconstructs samples by solving ODEs. The denoising objective is
\begin{equation}
    \mathbb{E}_{\bar{\bm{a}}\sim \mu}\mathbb{E}_{\bm{n}\sim \mathcal{N}_\sigma}[\lambda(\sigma)\|D_\theta(\bar{\bm{a}} + \bm{n}, \bar{\bm{o}}, \sigma) - \bar{\bm{a}}\|_2^2]
\end{equation}
where $D_\theta$ is the denoiser network. To handle padded dimensions, we introduce masked denoising score matching to ensure that the model focuses only on valid data entries. Here, $\odot$ denotes element-wise multiplication (Hadamard product).
\begin{equation}
    \mathbb{E}_{\bar{\bm{a}}\sim \mu}\mathbb{E}_{\bm{n}\sim \mathcal{N}_\sigma}[\lambda(\sigma)\|\bm{b} \odot \big(D_\theta(\bm{b} \odot(\bar{\bm{a}} + \bm{n}), \bar{\bm{o}}, \sigma) - \bar{\bm{a}}\big)\|_2^2]
\end{equation}

\subsection{Enhancing Diffusion with Residual Policy and Multi-Critic RL}

Although diffusion models provide strong priors, they may not fully capture fine-grained dynamics or task-specific variations. To address this, we introduce a residual policy shared across all robots
\begin{equation}
    \bm{a} = \bar{\bm{a}}_{\text{prior}} + \Delta \bm{a}
\end{equation}
where $\bar{\bm{a}}_{\text{prior}}$ is sampled from the diffusion model, and $\Delta \bm{a}$ is predicted by a residual policy $\pi_\theta(\Delta \bm{a} | \bar{\bm{o}}, \bar{\bm{a}}_{\text{prior}})$ trained via PPO. 

To enable generalization across robot embodiments, we train a single residual policy shared by all robots, while adopting a multi-critic PPO framework~\citep{mysore2022multi,xu2023composite, huang2025learning, wang2025beamdojo} to resolve optimization conflicts. Specifically, we instantiate a separate critic $V^k_\phi$ for each robot type $k = 1, \dots, K$, each receiving its corresponding privileged state $\bm{s}^k_t$ as input. During training, the shared actor $\pi_\theta$ is optimized using the following loss:
\begin{equation}
    J(\pi_\theta) = \sum_{k=1}^{K} \mathbb{E}_t \left[  \bar{A}_t^k(\bm{s}_t^k, \bm{a}_t^k, \Delta \bm{a}_t^k) \log \pi_\theta(\Delta \bm{a}^k | \bar{\bm{o}}^k, \bar{\bm{a}}_{\text{prior}}^k) \right]
\end{equation}
where $\bar{A}_t^k$ denotes the normalized advantage from the $k$-th critic. 

Each robot's reward function includes (i) task-specific components (e.g., velocity tracking, stability), (ii) regularization terms (e.g., torque penalties), and (iii) a residual penalty $r_d(\Delta \bm{a}_t)$ that encourages the residual to remain close to the generative prior, reducing overcorrection. The complete reward formulation is detailed in Appendix B.

During training, we run multiple robots in parallel environments and alternate between diffusion sampling and residual optimization. Only the residual policy is updated online, while the diffusion model remains frozen. This design allows the residual to specialize to task-level feedback while leveraging the shared generative prior.

\section{Evaluation}
We conducted comprehensive experiments to validate two core hypotheses:
\begin{itemize}[leftmargin=12pt]
    % \vspace{-8pt}
    \item \textit{Cross-embodiment generalization:} Our unified policy framework outperforms robot-specific baselines in controlling diverse morphologies under identical RL hyperparameters.
    % \vspace{-12pt}
    \item \textit{Zero-shot sim2real transfer:} The proposed residual policy effectively bridges the sim2real gap for diffusion-based controllers.
\end{itemize}
% \vspace{-8pt}
Our ablation and real-world deployment tests address three critical research questions:
\begin{itemize}[leftmargin=12pt]
    \item Cross-embodiment knowledge emergence: Does cross-robot training on heterogeneous embodiments facilitate the emergence of shared locomotion skills that remain unattainable through single-embodiment learning paradigms?
    \item Performance superiority: Can our cross-robot policy outperform morphology-specific reinforcement learning baselines in locomotion tasks when trained under identical parameter constraints and environmental conditions?
    \item Sim2real transfer:  Does the residual policy effectively enable zero-shot transfer of the multi-embodiment diffusion model to physical hardware?
\end{itemize}

\subsection{Evaluation Setup}
Our evaluation framework encompasses four distinct robots to systematically validate cross-morphology generalization capabilities: \textbf{(1) Point-foot biped} - Minimalist design with underactuated dynamics, \textbf{(2) Wheel-actuated biped} - Hybrid locomotion combining leg and wheel modalities, \textbf{(3) Full-scale humanoid} - High-DoF system with complex whole-body coordination, and \textbf{(4) Quadruped} - Dynamic quadrupedal locomotion with multiple contact points. 

The multi-morphological platforms enable a comprehensive evaluation of our unified control framework through both simulation and real-world experiments.  These tests demonstrate the framework’s robustness and adaptability in managing embodied variations across different robotic morphologies. The subsequent sections detail our methodology and experimental results using these platforms.

\subsection{Ablation Study}
To demonstrate the significance of each module within the unified framework, we have designed the following ablation studies:

\begin{itemize}[leftmargin=12pt]
    \vspace{-0.2cm}
    \item \textbf{RL Baseline}: Morphology-specific policies trained via PPO, following implementation practices from \texttt{humanoid\_gym}~\citep{gu2024advancing} and \texttt{legged\_gym}~\citep{rudin2022learning}.
    
    \item \textbf{Single-Robot Diffusion Policy (SR-DP)}: EDM trained on individual robot demonstrations.
    
    \item \textbf{Cross-Robot Diffusion Policy (CR-DP)}: Our EDM variant with dynamic action masking, trained on aggregated multi-robot datasets.
    
    \item \textbf{CR-DP with Residual Adaptation (CR-DP+RA)}: Our full approach integrating masked diffusion pre-training on multi-robot datasets
  with subsequent PPO-based residual policy adaptation through shared network parameters.
\end{itemize}

To ensure an equitable comparison across four methodologies, we maintained identical RL parameters when training both CR-DP+RA and RL baseline. Our evaluation protocol employs policies trained via baseline RL to generate rollout data for diffusion policy (DP) training, enabling direct comparison of average returns and auxiliary metrics within a unified environmental setup.

% \begin{figure}[H]
%   \centering
%   \vspace{-0.1cm}
%   \hspace{-0.5cm}
%   \includegraphics[width=0.9\textwidth]{fig_corl2025/reward_comparison.pdf}
%   \vspace{-0.3cm}
%   \caption{\footnotesize{Average Return of 4 Robots: To ensure an equitable comparison across four methodologies, we maintained identical RL parameters when training both CR-DP+RA and RL baseline. Our evaluation protocol employs policies trained via baseline RL to generate rollout data for diffusion policy (DP) training, enabling direct comparison of average returns and auxiliary metrics within a unified environmental setup.}}
%   \vspace{-0.5cm}
%   \label{fig:reward_comparison}
% \end{figure}

\begin{figure}[ht]
  % \vspace{-0.2cm}
  \centering
  \begin{minipage}[t]{0.49\textwidth}
    \centering
    \includegraphics[width=\linewidth]{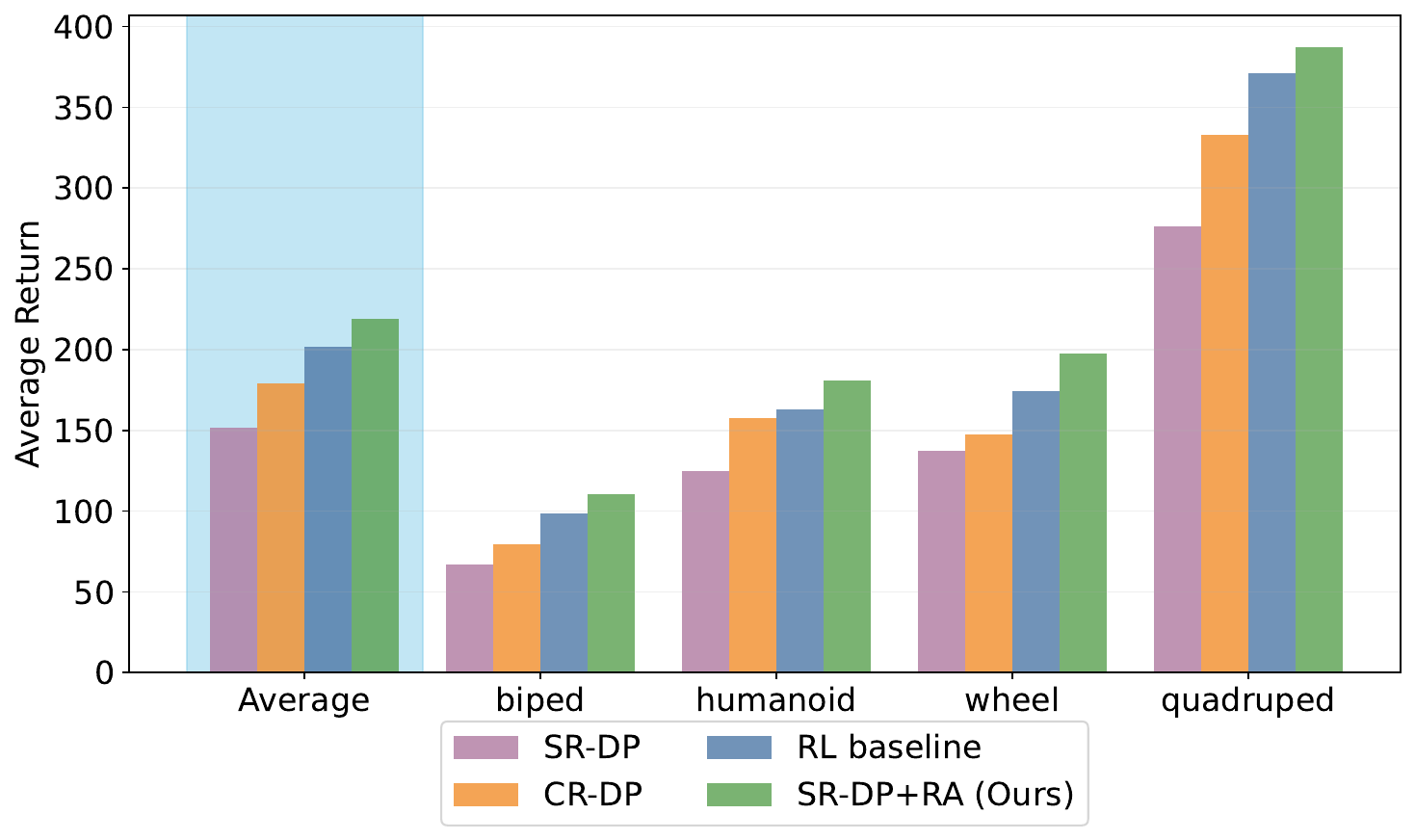}
    {\footnotesize{(a) Average Return of 4 Robots}}
    \label{fig:left}
  \end{minipage}
  \hfill
  \begin{minipage}[t]{0.49\textwidth}
    \centering
    \vspace{-4cm}
    \begin{minipage}[t]{\textwidth}
      \includegraphics[width=\linewidth]{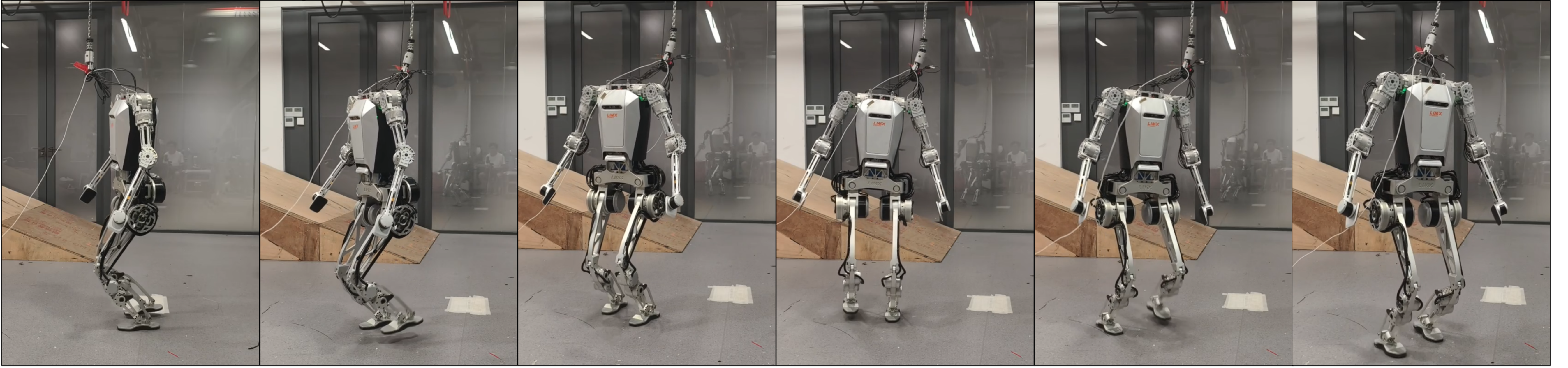}
      {\footnotesize{(b) Humanoid Locomotion Experiments}}
      \label{fig:right_top}
    \end{minipage}
    \begin{minipage}[t]{\textwidth}
      \includegraphics[width=\linewidth]{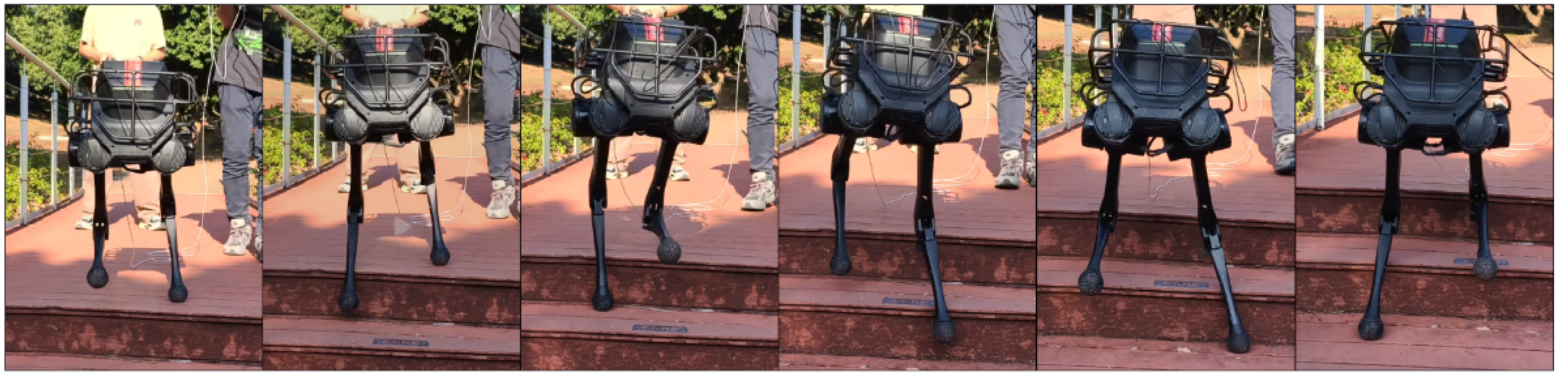}
      {\footnotesize{(c) Point-Foot Biped Downstairs}} 
      \label{fig:right_bottom}
    \end{minipage}
  \end{minipage}
  \caption{\footnotesize{(a) Comparative performance analysis of four robot morphologies (biped, humanoid, wheeled, quadruped) in terrain negotiation tasks. SR-DP+RA achieves 10.35\% average improvement over RL baseline (12.49\% biped, 13.57\% wheeled-biped, 4.38\% quadruped, 10.97\% humanoid), while CR-DP surpasses SR-DP by 17.96\% (17.81\% biped, 7.52\% wheeled-biped, 20.47\% quadruped, 26.02\% humanoid). (b) Humanoid locomotion demonstrates zero-shot sim2real transfer. (c) Point-foot biped successfully descends stairs}}
  \label{fig:reward_comparison}
\end{figure}

\clearpage
Our evaluation metrics are designed as follows:
\begin{itemize}[leftmargin=12pt]
\vspace{-0.2cm}
\item \textbf{(AR)} Average Return: Represents the cumulative RL return obtained by the agent during its survival period, up to a maximum of 20 seconds. It reflects the overall performance of the agent, incorporating both task success and control efficiency.
\item \textbf{(MEL)} Mean Episode Length: Represents the average survival duration of the agent in each episode, with a maximum of 20 seconds. A longer episode length indicates better stability.
\item \textbf{(LVT/AVT)} Mean Episode Linear/Angular Velocity Tracking Reward: Represents the reward associated with tracking a desired linear/angular velocity. 
% This reward is inversely proportional to the linear/angular velocity tracking error, following the formula  
% \[
% R_{\text{lin}} = \exp(-\alpha \cdot e_{\text{lin}}^2),
% \]  
% \[
% R_{\text{ang}} = \exp(-\beta \cdot e_{\text{ang}}^2),
% \]  
% where \( e_{\text{lin}} \) and \( e_{\text{ang}} \) are the linear and angular velocity error and \( \alpha \) and \( \beta \) are scaling factor. Higher rewards indicate better velocity tracking accuracy.
\end{itemize}

% \begin{wrapfigure}{r}{0.54\textwidth} 
%   \centering
%   \vspace{-0.1cm}
%   \hspace{-0.5cm}
%   \includegraphics[width=0.52\textwidth]{fig_corl2025/reward_comparison.pdf}
%   \vspace{-0.3cm}
%   \caption{\footnotesize{Average Return of 4 Robots}}
%   \vspace{-0.5cm}
%   \label{fig:reward_comparison}
% \end{wrapfigure}

\begin{table}[htbp]
\scriptsize
\vspace{-0.5cm}
\centering
\caption{\footnotesize{Cross-Morphology Performance Comparison of Ablation Studies}}
\label{tab:cross_ablation}
\begin{tabular}{lcccccc}
\toprule
\multirow{2}{*}{\textbf{Method}} 
& \multicolumn{3}{c}{\textbf{Point-Foot}} 
& \multicolumn{3}{c}{\textbf{Wheeled}} \\
\cmidrule(lr){2-4} \cmidrule(lr){5-7}
& MEL$\uparrow$ & LVT$\uparrow$ & AVT$\uparrow$ 
& MEL$\uparrow$ & LVT$\uparrow$ & AVT$\uparrow$ \\
\midrule
Baseline-point-foot
& $\mathbf{18.71\pm0.37}$ & $1.39\pm0.25$ & $0.94\pm0.17$ 
& - & - & - \\
SR-DP-point-foot 
& $14.68\pm0.66$ & $0.93\pm0.37$ & $0.71\pm0.26$ 
& - & - & - \\
Baseline-wheeled
& - & - & - 
& $18.29\pm0.37$ & $2.48\pm0.58$ & $1.28\pm0.30$ \\
SR-DP-wheeled 
& - & - & - 
& $18.63\pm0.33$ & $2.37\pm0.55$ & $1.26\pm0.28$ \\
\cmidrule{1-7}
CR-DP (Ours) 
& $17.28\pm0.56$ & $1.15\pm0.31$ & $0.87\pm0.21$ 
& $\mathbf{18.97\pm0.35}$ & $2.54\pm0.50$ & $1.35\pm0.26$ \\
CR-DP+RA (Ours) 
& $18.13\pm0.46$ & $\mathbf{1.42\pm0.29}$ & $\mathbf{0.97\pm0.20}$ 
& $18.86\pm0.33$ & $\mathbf{2.72\pm0.53}$ & $\mathbf{1.36\pm0.28}$ \\
\midrule
\midrule
\multirow{2}{*}{\textbf{Method}} 
& \multicolumn{3}{c}{\textbf{Humanoid}} 
& \multicolumn{3}{c}{\textbf{Quadruped}} 
\\\cmidrule(lr){2-4} \cmidrule(lr){5-7}
& MEL$\uparrow$ & LVT$\uparrow$ & AVT$\uparrow$ 
& MEL$\uparrow$ & LVT$\uparrow$ & AVT$\uparrow$ \\
\midrule
Baseline-humanoid
& $18.88\pm0.43$ & $1.57\pm0.27$ & $1.23\pm0.20$ 
& - & - & - \\
SR-DP-humanoid
& $13.84\pm0.78$ & $1.04\pm0.42$ & $0.93\pm0.34$
& - & - & - \\
Baseline-quadruped
& - & - & - 
& $\mathbf{19.79\pm0.17}$ & $\mathbf{5.45\pm0.37}$ & $\mathbf{4.34\pm0.30}$ \\
SR-DP-quadruped
& - & - & - 
& $16.90\pm0.62$ & $4.51\pm1.25$ & $3.32\pm0.95$ \\
\cmidrule{1-7}
CR-DP (Ours) 
& $17.13\pm0.54$ & $1.33\pm0.36$ & $1.17\pm0.29$ & $19.31\pm0.29$ & $5.29\pm0.57$ & $4.19\pm0.47$ \\
CR-DP+RA (Ours) 
& $\mathbf{19.11\pm0.44}$ & $\mathbf{1.58\pm0.28}$ & $\mathbf{1.35\pm0.23}$& $19.60\pm0.19$ & $5.42\pm0.50$ & $4.32\pm0.41$ \\
\bottomrule
\end{tabular}
\vspace{-10px}
\end{table}

\subsubsection{Emergence of Shared Locomotion Skills via Cross-Embodiment Learning} 
\begin{wrapfigure}{r}{0.6\textwidth} 
  \centering
  \vspace{-0.4cm}
  \hspace{-0.5cm}
  \includegraphics[width=0.6\textwidth]{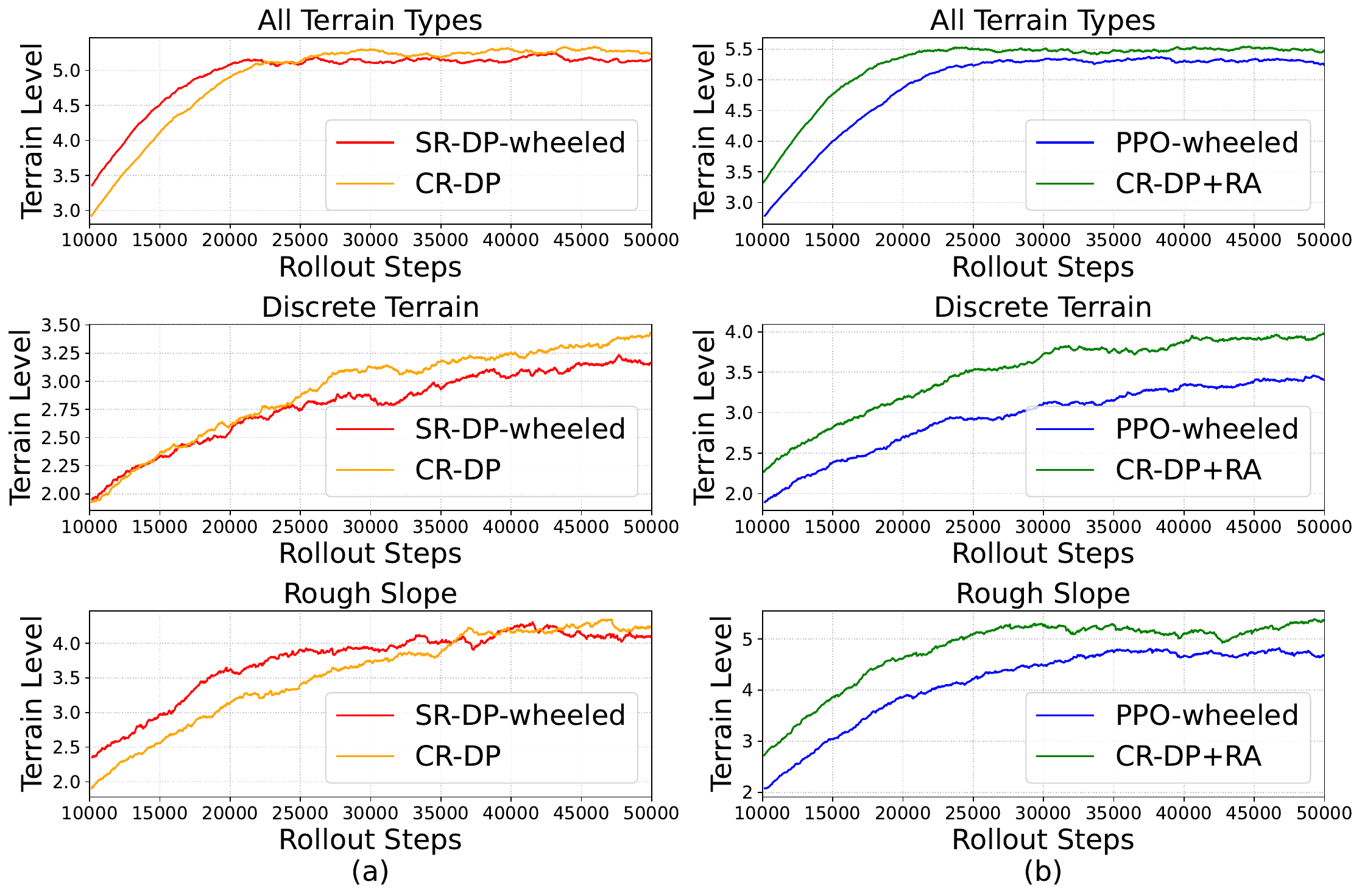}
  \vspace{-0.3cm}
  \caption{\footnotesize{Terrain traversal performance of wheeled-biped robots under different training setups.  
(a) Performance of the diffusion model trained using offline datasets only (SR-DP-Wheeled vs. CR-DP) on terrains seen during training.
(b) Comparison between PPO-trained baseline and our diffusion+residual policy (CR-DP+RA), which incorporates online reinforcement learning.  
CR-DP shows improved adaptability over SR-DP-Wheeled on rougher terrains, while CR-DP+RA further enhances robustness and outperforms PPO in terms of stability and obstacle traversal.}}
  \vspace{-0.5cm}
  \label{fig:Terrain Level}
\end{wrapfigure}

Emergence of shared locomotion skills via cross-embodiment learning refers to the phenomenon where robotic agents with distinct morphologies learn fundamental movement patterns through shared learning frameworks. 

As evidenced in Figure \ref{fig:Terrain Level}, our proposed CR-DP achieves an overall terrain traversal improvement of 2.67\% over SR-DP for wheeled-bipeds, with particularly surges in challenging terrains: 6.10\% improvement over discrete obstacles and 3.87\% on rough slopes, demonstrating its adaptability to complex environments.

% \begin{figure}[htbp]
%   \centering 
  
%   \begin{minipage}[b]{0.95\textwidth}
%     \centering
%     \includegraphics[width=\textwidth]{fig_corl2025/terrain_level_group1.pdf}
%     \caption*{(a)\footnotesize{Dataset-Specific Diffusion Policy Terrain Adaptation: Comparative Analysis of Diffusion Models Trained on Wheeled (SR-DP-wheeled) vs. Cross-Robot Datasets (CR-DP)}}
%     \label{fig:sub1}
%   \end{minipage}
  
%   % \vspace{0.5cm}
  
%   \begin{minipage}[b]{0.95\textwidth}
%     \centering
%     \includegraphics[width=\textwidth]{fig_corl2025/terrain_level_group2.pdf}
%     \caption*{(b)\footnotesize{RL Policy Comparison: Standard RL Approach (RL baseline-wheeled) vs. Diffusion Pretraining with Residual Policy (CR-DP+RA) Framework}}
%     \label{fig:sub2}
%   \end{minipage}

%   \caption{\footnotesize{Comparative Terrain Negotiation of Wheeled-Biped Robots: SR-DP-Wheeled, CR-DP, and CR-DP+RA rollout in
% training terrain composition: Flat (30\%), smooth slopes (30\%), rough slopes (10\%), descending stairs (10\%), discrete obstacles (20\%).
% Performance summary: CR-DP slightly exceeds terrain adaptability over SR-DP-Wheeled across rough slopes and discrete obstacles, while maintaining parity on flat/smooth terrains. CR-DP+RA further enhances robustness in complex environments, surpassing RL baseline performance in both stability and obstacle clearance efficiency.}}
%   \label{fig:Terrain Level}
%   \vspace{-10px}
% \end{figure}

Notably, under identical parameter configurations, the CR-DP with Residual Adaptation variant demonstrates measurable performance gains over the RL Baseline in wheeled-biped scenarios, achieving a 15.13\% improvement on discrete terrain, 11.57\% enhancement on rough slopes, and an overall terrain navigation improvement of 3.46\%.
This is particularly remarkable given that the original wheeled-biped robot dataset and RL training parameters failed to enable leg-lifting locomotion - which fundamentally restricted mobility in discontinuous terrains such as stepped surfaces and stair-like obstacles. Our experiments demonstrate that the cross-robot DP component effectively transfers terrain negotiation expertise from heterogeneous embodiments to the wheeled-biped platform, enabling stable traversal of challenging terrains.

Figure \ref{fig:reward_comparison} reveals that CR-DP achieves comprehensively higher average returns than SR-DP, while CR-DP with Residual Adaptation substantially exceeds RL Baseline performance, which achieves a 10.35\% improvement on average, with gains up to 13.57\% in wheeled-biped locomotion tasks. These results strongly suggest that the cross-robot component successfully captures shared locomotion skills.
An interesting observation emerges in quadruped scenarios: CR-DP with Residual Adaptation matches RL Baseline performance in MEL and LVT/AVT Reward metrics. We hypothesize this reflects asymmetric knowledge transfer - quadruped locomotion experiences provide exemplary guidance for other morphologies, while motion patterns from other embodiments offer limited benefit to quadruped agents.

\subsubsection{Multi-Loco Policy Outperforms Morphology-Specific RL Baseline}
As quantitatively demonstrated in Figure \ref{fig:reward_comparison} and Table \ref{tab:cross_ablation}, our CR-DP with Residual Adaptation framework achieves consistent performance advantages over the RL Baseline across all evaluation metrics and four robotic embodiments.
The supervised learning foundation of CR-DP reaches 87.78\% of the baseline RL's maximum average return potential, consistent with known limitations of behavioral cloning paradigms~\citep{ross2010efficient}.
Through integration with Residual Adaptation - implemented using identical RL hyperparameters - the hybrid architecture demonstrates significant performance improvements, attaining 113.57\% of baseline effectiveness.
This performance inversion reveals the critical role of our hybrid learning framework: while the diffusion model component distills transferable motion skills from heterogeneous robot experiences, the residual RL module dynamically adapts these skills to embodiment-specific dynamics through online RL.

\subsubsection{Zero-Shot sim2real Transfer: Experiments Evaluation for the Robust Locomotion} 
We implemented the unified locomotion policy across all four robots (embodiment details in Fig.\ref{fig:framework}), with real-time control executed on an Intel i9-13900HX CPU at 50Hz refresh rates using EtherCAT communication.

As demonstrated in Fig.\ref{fig:real world experiments} and Fig.\ref{fig:reward_comparison}(b)(c), our Multi-Loco policy enables robust locomotion capabilities across challenging environments. The framework facilitates smooth embodiment transitions and stable gait generation in diverse terrain geometries ranging from inclined surfaces to vegetated landscapes and irregular substrates. This operational consistency across heterogeneous environmental conditions confirms the policy's capacity to maintain dynamic stability through adapting its motor patterns, highlighting its practical viability for deployment in unstructured real-world settings.

\section{Conclusions}
In this work, we introduced Multi-Loco, a unified framework for multi-embodiment legged locomotion that integrates generative diffusion models with reinforcement learning. Our approach successfully addresses the challenge of generalizing control policies across diverse robot embodiments, enabling seamless adaptation to varying hardware configurations. 

Experimental results across simulation and hardware platforms demonstrate the framework's ability to generalize effectively, highlighting its potential for real-world applications in dynamic and unpredictable environments. By leveraging masked diffusion model, Multi-Loco gracefully handles differences in observation and action spaces, ensuring scalability across diverse robotic systems.

In conclusion, Multi-Loco pioneers a novel pathway for tackling the challenges of multi-embodiment unified control policy, delivering a scalable and flexible framework that enables robust and adaptive locomotion across diverse robotic embodiments.

\clearpage
\section{Limitations and Future Works}
The proposed Multi-Loco framework demonstrates strong performance in unifying locomotion policies for diverse embodiments, including point-foot biped, wheeled biped, quadruped and humanoid, marking a foundational step toward general-purpose control policy synthesis for multi-embodiment robotic systems. However, several limitations remain, which we discuss below alongside potential future directions.

A key limitation of the current framework is its reliance on observation-action paired datasets for training. This restricts its ability to leverage widely available motion capture data (e.g., human or animal locomotion), which often lack explicit action labels but encode rich motor skills. Future work will focus on action-free training paradigms to integrate such motion data, enabling the framework to exploit larger and more diverse datasets.

While Multi-Loco successfully unifies policies for predefined embodiments, direct synthesis of locomotion policies for entirely new morphologies remains an open challenge. Future efforts will investigate how shared representations of locomotion principles, learned from existing embodiments, can be extended to novel morphologies through zero-shot adaptation or few-shot fine-tuning, reducing the need for extensive retraining.

By addressing these challenges, we aim to advance Multi-Loco into a universal locomotion synthesis framework capable of generalizing across embodiments, tasks, and real-world conditions.

%===============================================================================
% \clearpage
% % The acknowledgments are automatically included only in the final and preprint versions of the paper.
% \acknowledgments{If a paper is accepted, the final camera-ready version will (and probably should) include acknowledgments. All acknowledgments go at the end of the paper, including thanks to reviewers who gave useful comments, to colleagues who contributed to the ideas, and to funding agencies and corporate sponsors that provided financial support.}

%===============================================================================

% no \bibliographystyle is required, since the corl style is automatically used.
\bibliography{ref}  % .bib
\clearpage
\appendix
\section*{Appendix}
\section*{Experiment Videos}
We conducted comprehensive real-world evaluations of our framework across four distinct legged robotic platforms. For detailed empirical validation, we encourage readers to view the supplementary video. As demonstrated in the experimental recordings, our framework demonstrates the capability of a unified control policy to govern four morphologically diverse legged robots with different actuator configurations and mass distributions. This cross-platform adaptability is achieved through our novel methodology, which enables robust policy generalization while maintaining dynamic locomotion performance.
\section{Details of Diffusion Model Training and Inference}
DDPM, as one kind of diffusion models, has achieved remarkable results in tasks such as robotic manipulation and locomotion for legged robots due to its ability to represent multimodality and complex distributions. Due to the high number of denoising steps required by DDPM, the inference time can be quite long. For humanoid robots, the diffusion policy must operate at a frequency of at least 50 Hz, ideally reaching 100 Hz. Therefore, it is necessary to either accelerate its performance or replace it with alternatives such as DDIM, EDM, or consistency models. In this context, we have chosen the EDM model to ensure high sample quality while minimizing the inference time to enable high-frequency feedback control.

EDM (Elucidated Diffusion Model)~\citep{edm} is based on the reverse-time ODE of variance-exploding SMLD~\citep{song2020score}. In order to elucidate the design space of diffusion model, EDM proposes a more general form while considering time-dependent scaling and introduces preconditioning to cancel the effects caused by increasing noise variance in denoising score matching. EDM reparametrized the denoiser $D_\theta(\hat\bx,\sigma)$ as the following form
\begin{equation}
    D_\theta(\hat\bx,\sigma) = c_{\text{skip}}(\sigma)\hat\bx + c_{\text{out}}(\sigma)F_\theta(c_{\text{in}}(\sigma)\hat\bx,c_{\text{noise}}(\sigma)).
\end{equation}
and the corresponding objective of denoising score matching is changed to
\begin{equation}
    \mathbb{E}_{\bx,\bn}\left[\frac{c_{\text{out}}(\sigma)^2}{\sigma}\|F_\theta(c_{\text{in}}(\sigma)(\bx+\bn),c_{\text{noise}}(\sigma)) - \bx_d\|_2^2\right]
\end{equation}
where $\bx_d = (\bx-c_{\text{skip}}(\sigma)(\bx+\bn))/c_{\text{out}}$. With the requirements of unit variance of input and output while minimizing $c_{\text{out}}$, the value of parameters are chosen as below:
\begin{subequations}
    \begin{align}
        c_{\text{noise}}(\sigma) &= \ln(\sigma)/4,\\
        c_{\text{in}}(\sigma) &= 1/ \sqrt{\sigma_{\bx}^2 + \sigma^2}, \\
        c_{\text{skip}}(\sigma) &= \sigma^2_{\bx}/(\sigma^2 + \sigma^2_{\bx}),\\
        c_{\text{out}}(\sigma) &= \sigma\cdot\sigma_{\bx}/(\sigma^2 + \sigma^2_{\bx}),
    \end{align}
\end{subequations}
where $\sigma_{\bx}$ is the variance of the data. \citep{edm} suggests that the optimal choice for this function is $\sigma(t)=t$, an approach we also adopt in our work. As a result, EDM solves the following probability flow ODE 
\begin{equation}
    \frac{\text{d}\bx}{\text{d}\bar{t}} = \frac{\bx - D_\theta(\bx,\sigma)}{t}
\end{equation}
for sampling and starting from $\bx(T)\sim \N(0,T^2\bm{I})$ and stopping at $\bx(0)$. In practice, we use Euler method to solve this ODE for sampling.

\begin{figure}[htbp!]
    \centering
    \includegraphics[width=0.65\linewidth]{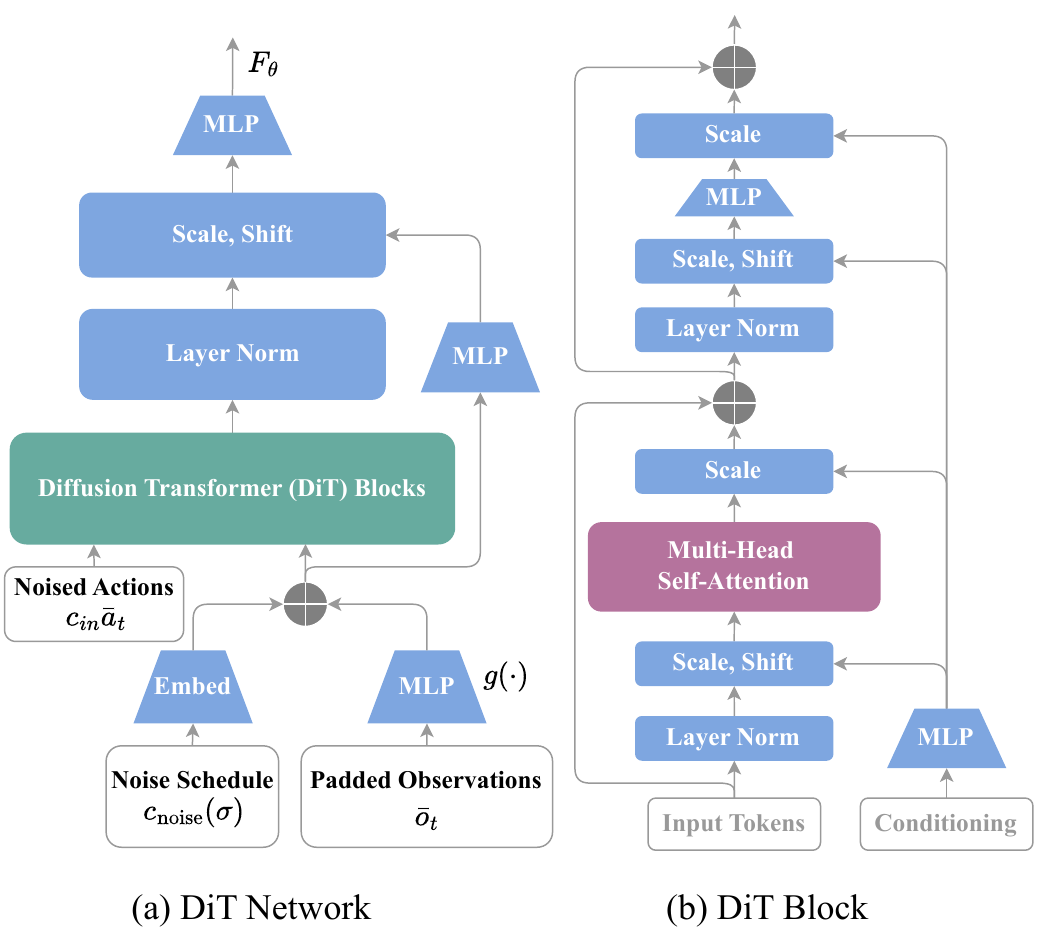}
    \caption{Neural network structure of DiT which is used to fit $F_\theta$.}
    \label{fig: dit}
\end{figure}

In terms of network architecture, diffusion models commonly utilize U-Net and Transformer-based structures. In our scenario, we chose DiT (Denoising Transformer) model~\citep{dit} as backbone to fit $F_\theta(\cdot)$. The padded observations $\bar{\bm{o}}_t$ as condition variable was embed into a latent space by a multilayer perceptron (MLP) denoted by $g(\cdot)$. To incorporate $\bar{\bm{o}}_t$  as condition variable, $D_\theta(\hat\bx,\sigma)$ is rewriten as $D_\theta(\hat\ba_t,\bar{\bm{o}}_t,\sigma)$ and $F_\theta(c_{\text{in}}\hat\bx,c_{\text{noise}}(\sigma))$ is changed to $F_\theta(c_{\text{in}}\hat\ba_t, g(\bar{\bm{o}}_t),c_{\text{noise}}(\sigma))$. The details of the network structure can be found in Fig.~\ref{fig: dit}.

\begin{table}[htbp!]
\footnotesize
\renewcommand{\arraystretch}{1.2} 
\centering
\begin{tabular}{l|l}
\hline
Parameter & Value / Description \\
\hline
\multicolumn{2}{c}{Data Samples} \\
\hline
Prediction Horizon & 4 \\
Observation Horizon & 10 \\
Maximum Samples per Robot & 2048000  \\
Batch Size & 8192 \\
\hline
\multicolumn{2}{c}{Diffusion Network (DiT)} \\
\hline
Input Dimension & 20 \\
Embedding Dimension (emb\_dim) & 128 \\
Model Dimension (d\_model) & 256 \\
Number of Attention Heads (n\_heads) & 8 \\
Network Depth (depth) & 3 \\
Timestep Embedding Type & Fourier \\
\hline
\multicolumn{2}{c}{MLP Condition Network $g(\cdot)$} \\
\hline
Input Dimension & 683 \\
Output Dimension & emb\_dim \\
Hidden Dimensions (hidden\_dim) & [512, 256] \\
Activation Function & ELU \\
\hline
\multicolumn{2}{c}{Training Parameters} \\
\hline
Learning Rate & 3e-4 \\
Optimizer & Adam \\
Learning Rate Scheduler & Cosine Annealing \\
EMA Rate & 0.999 \\
Number of Epochs & 500 \\
Seed & 3407 (inspired by ~\citep{picard2021torch}) \\
\hline
\end{tabular}
\vspace{0.3cm}
\caption{Configurations for Unified Diffusion Model}
\label{tab: param}
\end{table}

\begin{algorithm}
\caption{Masked EDM Training and Inference}
\begin{algorithmic}[1]
\State \textbf{Initialize:} 
\State - Networks: DiT $F_\theta$ and Condition MLP $g_\phi$
\State - EDM params: $\sigma_\text{data}=0.5$, $\sigma_\text{min}=0.002$, $\sigma_\text{max}=80.0$, $\rho=7.0$
\State - Noise sampling: $P_\text{mean}=-1.2$, $P_\text{std}=1.2$
\State - Training: batch size $=512$, learning rate $=3\times10^{-4}$, EMA rate $=0.999$

\Procedure{DataPreprocessing}{}
    \State Extract trajectory segments (observations $\bm{o}$, actions $\bm{a}$)
    \State Apply zero-padding to handle variable-length sequences
    \State Compute quantile-based MinMax normalization:
    \State - Calculate 5\% and 95\% quantiles for each feature dimension
    \State - Normalize to [-1,1] range: $\bm{x}_\text{norm} = 2 \cdot \frac{\bm{x} - \bm{x}_\text{min}}{\bm{x}_\text{max} - \bm{x}_\text{min}} - 1$
    \State Create condition vector by concatenating normalized obs and commands
    \State \Return Normalized dataset, normalization statistics
\EndProcedure

\Procedure{Train}{Dataset $\mathcal{D}$, Epochs $=400$K}
    \For{each epoch}
        \For{batch $(\bm{o}, \bm{a}_0, \bm{b}) \sim \mathcal{D}$}
            \State Sample noise $\sigma \sim e^{\mathcal{N}(-1.2, 1.2^2)}$, $\epsilon \sim \mathcal{N}(0, I)$
            \State Compute noisy actions $\bm{a}_t = \bm{a}_0 + \sigma \cdot \epsilon$
            \State Compute EDM preconditioning coefficients:
            \State $c_\text{skip} = \frac{0.5^2}{0.5^2 + \sigma^2}$
            \State $c_\text{out} = \frac{\sigma \cdot 0.5}{\sqrt{0.5^2 + \sigma^2}}$
            \State $c_\text{in} = \frac{1}{\sqrt{0.5^2 + \sigma^2}}$
            \State $c_\text{noise} = 0.25 \cdot \log \sigma$
            \State Compute network prediction: $D_\theta(\bm{a}_t, \sigma, g_\phi(\bm{o}))$
            \State Compute masked loss: $\mathcal{L} = ((1 - \bm{b}) \cdot (D_\theta(\bm{a}_t, \sigma, g_\phi(\bm{o})) - \bm{a}_0))^2$
            \State Apply EDM weighting: $\mathcal{L}_\text{final} = \left(\mathcal{L} \cdot \frac{0.5^2 + \sigma^2}{(\sigma \cdot 0.5)^2}\right).mean()$
            \State Update parameters using Adam ($\eta=3\times10^{-4}$)
            \State Update EMA model parameters (rate $=0.999$)
        \EndFor
    \EndFor
\EndProcedure

\Procedure{Inference}{Observation $\bm{o}$, Action mask $\bm{b}$, Sampling steps $S=5$}
    \State Normalize observations using stored statistics
    \State Initialize $\bm{a}_S \sim \mathcal{N}(0, 80.0^2 \cdot I)$
    \State Compute noise schedule: $\sigma_i = \left(0.002^{1/7} + \frac{i}{S}(80.0^{1/7} - 0.002^{1/7})\right)^7$ for $i \in \{0,1,...,S\}$
    \For{$i = S$ down to $1$}
        \State Apply action mask: $\bm{a}'_i = (1 - \bm{b}) \cdot \bm{a}_i$
        \State Compute network prediction with preconditioning
        \State ODE update: $\dot{\bm{a}} = \frac{\bm{a}_i - D_\theta(\bm{a}'_i, \sigma_i, g_\phi(\bm{o}))}{\sigma_i}$
        \State Euler step: $\bm{a}_{i-1} = \bm{a}_i - \dot{\bm{a}} \cdot (\sigma_i - \sigma_{i-1})$
    \EndFor
    \State Denormalize using stored statistics
    \State \Return $\bm{a}_\text{final}$
\EndProcedure
\end{algorithmic}
\label{alg: edm}
\end{algorithm}

\subsection{Hyperparameters and Training Hardware}
The DiT model configuration and training parameters are listed in Table.~\ref{tab: param}. Our implementation builds on the open-source \textit{CleanDiffuser} library~\cite{cleandiffuser}, which provides a modular and extensible interface for diffusion models in decision-making tasks. We adopt its implementation of EDM sampling for both training and inference. If the model is trained on a single NVIDIA 3090 GPU, it costs total training time ranging from 3 to 12 hours depending on robot embodiment and dataset size.

\subsection{Details of Training and Inference}
Our diffusion-based policy handles diverse embodiments through a unified framework that accommodates variable-dimensional observation and action spaces. During training, we normalize all inputs using quantile-based MinMax normalization with 5th and 95th percentiles to mitigate the effects of outliers. This maps features to a common [-1, 1] range while preserving the relative scaling of the majority of data points.

To support cross-embodiment generalization, we zero-pad observation and action vectors to match the largest dimension across all robot morphologies (observations: 68D, actions: 20D) and the dimension configurations of . During both training and inference, we apply morphology-specific binary masks $\bm{b}$ to ensure only valid dimensions contribute to the loss computation and prediction. Specifically, during training, the loss is computed as:
$\mathcal{L} = ((1 - \bm{b}) \cdot (D_\theta(\bm{a}_t, \sigma, g_\phi(\bm{o})) - \bm{a}_0))^2$

During inference, we employ the Euler method with 5 sampling steps to solve the ODE and generate actions. At each denoising step, we apply the action mask to ensure prediction consistency: $\bm{a}'_i = (1 - \bm{b}) \cdot \bm{a}_i$. 

\begin{table}[h]
\centering
\scriptsize
\begin{tabular}{ccccc}
\toprule
\textbf{Robot} & \textbf{Observation} & \textbf{Action} & \textbf{Command} & \textbf{Components} \\
\midrule
Point-Foot Biped & 
$\mathcal{O}_1 \in \mathbb{R}^{26}$ & 
$\mathcal{A}_1 \in \mathbb{R}^6$ &
$\mathcal{C}_1 \in \mathbb{R}^3$ &
Base pose (7D), velocity (6D), joint states (6D), gait phase (7D) \\
\addlinespace

Wheeled Biped & 
$\mathcal{O}_2 \in \mathbb{R}^{28}$ & 
$\mathcal{A}_2 \in \mathbb{R}^8$ &
$\mathcal{C}_2 \in \mathbb{R}^3$ &
Base pose (7D), velocity (6D), joint states (8D), wheel states (7D) \\
\addlinespace

Humanoid & 
$\mathcal{O}_3 \in \mathbb{R}^{68}$ & 
$\mathcal{A}_3 \in \mathbb{R}^{20}$ &
$\mathcal{C}_3 \in \mathbb{R}^3$ &
Base pose (7D), velocity (6D), joint states (48D), gait phase (7D) \\
\addlinespace

Quadruped & 
$\mathcal{O}_4 \in \mathbb{R}^{44}$ & 
$\mathcal{A}_4 \in \mathbb{R}^{12}$ &
$\mathcal{C}_4 \in \mathbb{R}^3$ &
Base pose (7D), velocity (6D), joint states (24D), gait phase (7D) \\
\bottomrule
\end{tabular}
\vspace{0.2cm}
\caption{\footnotesize{Configuration-Specific Observation and Action Spaces}}
\label{tab: config_spaces}
\end{table}

\subsection{Ablation Study: Diffusion Sampling Steps}
\begin{figure}[h]
    \centering
    \includegraphics[width=1.0\linewidth]{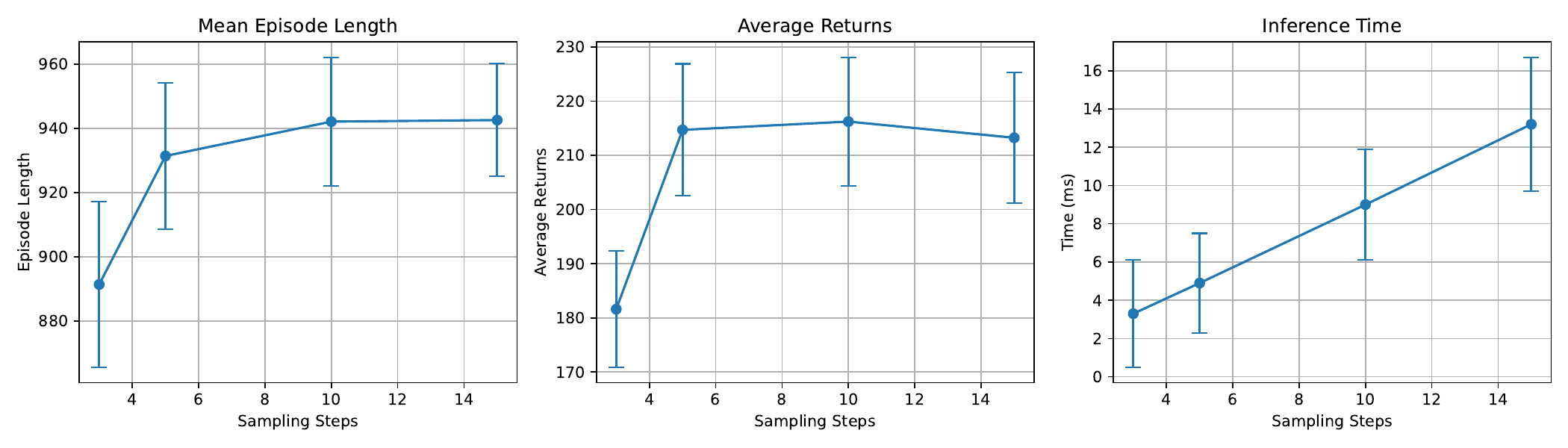}
    \caption{Comparison with Different Sampling Steps in EDM}
    \label{fig: sample_steps}
\end{figure}

We evaluate how the number of diffusion sampling steps impacts control performance. Fewer steps reduce computation time but may degrade trajectory quality. Results (Fig~\ref{fig: sample_steps}) show that using [3, 5,10,15] denoising steps balances control accuracy and sampling efficiency, with diminishing returns beyond 20 steps. According to this results, we choose 5 denoising steps in the deployment experiments.

\subsection{Ablation Study: Dataset Size for Diffusion Training}
To assess data efficiency, we train diffusion models on varying dataset sizes (1\%, 10\%, 25\%, 50\%, 100\%) of biped robot while keep others unchagned. The performance scales sublinearly with data, suggesting strong generalization even with partial data. However, extremely small datasets ($<25\%$) lead to unstable behaviors and poor terrain negotiation.

\begin{figure}[h]
\vspace{-0.5cm}
    \centering
    \includegraphics[width=1.0\linewidth]{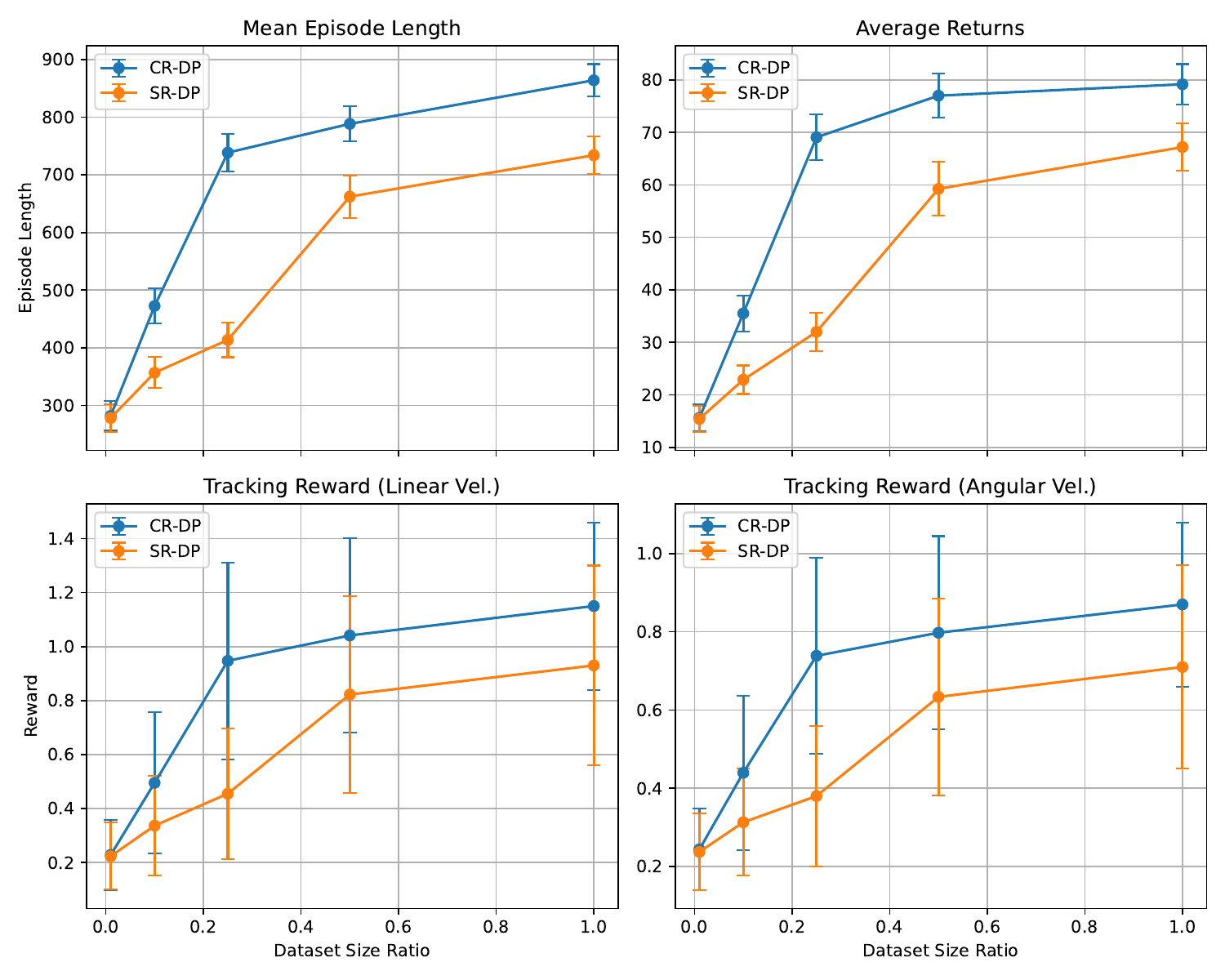}
    \caption{Biped Statistics: Comparison with EDMs Trained with Different Datasets Size. (The maximum number of samples of bipedal robots in the dataset is 2048000)}
    \label{fig: dataset_size}
    \vspace{-0.5cm}
\end{figure}

\subsection{Ablation Study: Dataset Composition Analysis for Diffusion Training}
\begin{figure}[h]
\vspace{-0.5cm}
    \centering
    \includegraphics[width=1.0\linewidth]{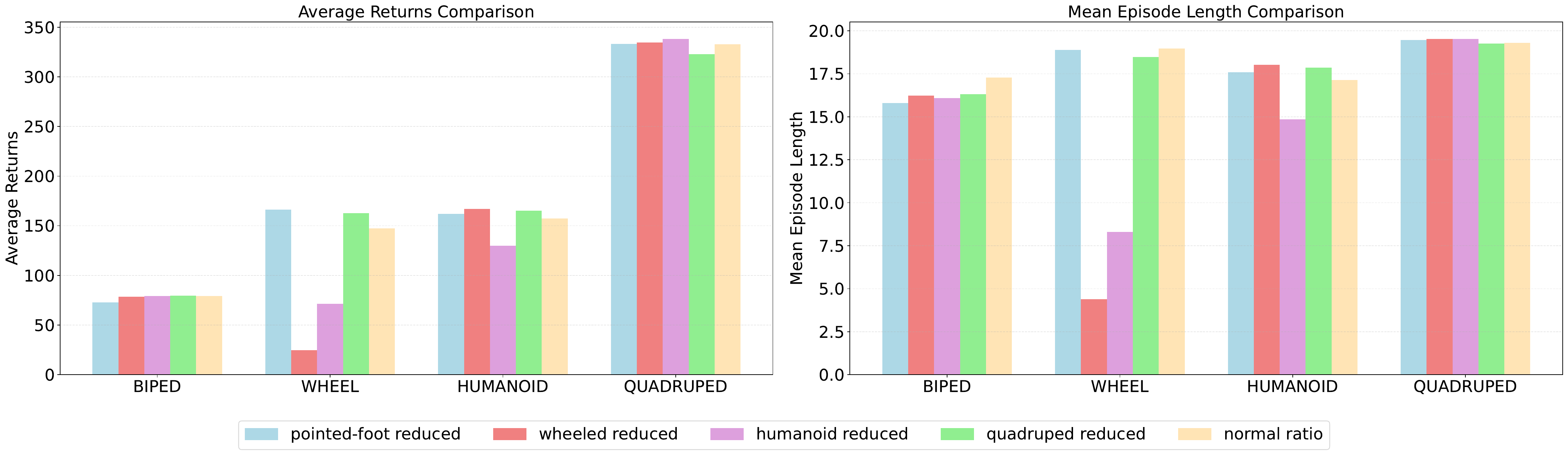}
    \caption{Impact of dataset composition ratios on diffusion policy training. The baseline normal ratio configuration (25\%-25\%-25\%-25\%) equally distributes data across four morphologies: pointed-foot biped, wheeled biped, humanoid, and quadruped. Four experimental variations were created by reducing one morphology's share to 10\% while proportionally increasing others to 30\% each: Pointed-foot reduced (10\%-30\%-30\%-30\%), Wheeled reduced (30\%-10\%-30\%-30\%), Humanoid reduced (30\%-30\%-10\%-30\%), and Quadruped reduced (30\%-30\%-30\%-10\%).}
    \label{fig: datasize_ratio_comparison}
\end{figure}

\vspace{-0.2cm}
We conducted a systematic investigation into how varying data distribution ratios across robot morphologies impact policy performance while keeping the total dataset size constant. Using a baseline configuration with equal 25\%-25\%-25\%-25\% allocations for four robot types (pointed-foot biped, wheeled biped, humanoid, and quadruped), we created four experimental conditions by sequentially reducing one category to 10\% while proportionally increasing others to 30\%. The resulting Average Return (AR) and Mean Episode Length (MEL) metrics were subsequently analyzed in Fig \ref{fig: datasize_ratio_comparison}.

Reducing data allocation for pointed-foot bipeds and quadrupeds showed negligible performance impacts, suggesting their relatively simple locomotion tasks require minimal training data. Conversely, decreasing wheeled biped data caused significant performance degradation in its corresponding policy, likely due to the inherent complexity of wheeled locomotion combined with limited cross-morphology knowledge transfer from more dissimilar robot types. Most notably, humanoid data reduction demonstrated dual impacts - not only expected self-performance deterioration from reduced training on its high-dimensional observation space, but also an unexpected AR decrease in wheeled biped performance. This cross-domain dependency confirms the hypothesis presented in our main text regarding humanoid data's complementary role in enhancing wheeled locomotion capabilities, suggesting previously unrecognized synergies between morphologically distinct systems.

\subsection{Zero-Shot Transfer to an Unseen Platform: Unitree Go2}
\vspace{-0.2cm}

\begin{wrapfigure}{l}{0.5\textwidth} 
    \vspace{-0.3cm}
    \centering
    \includegraphics[width=1.0\linewidth]{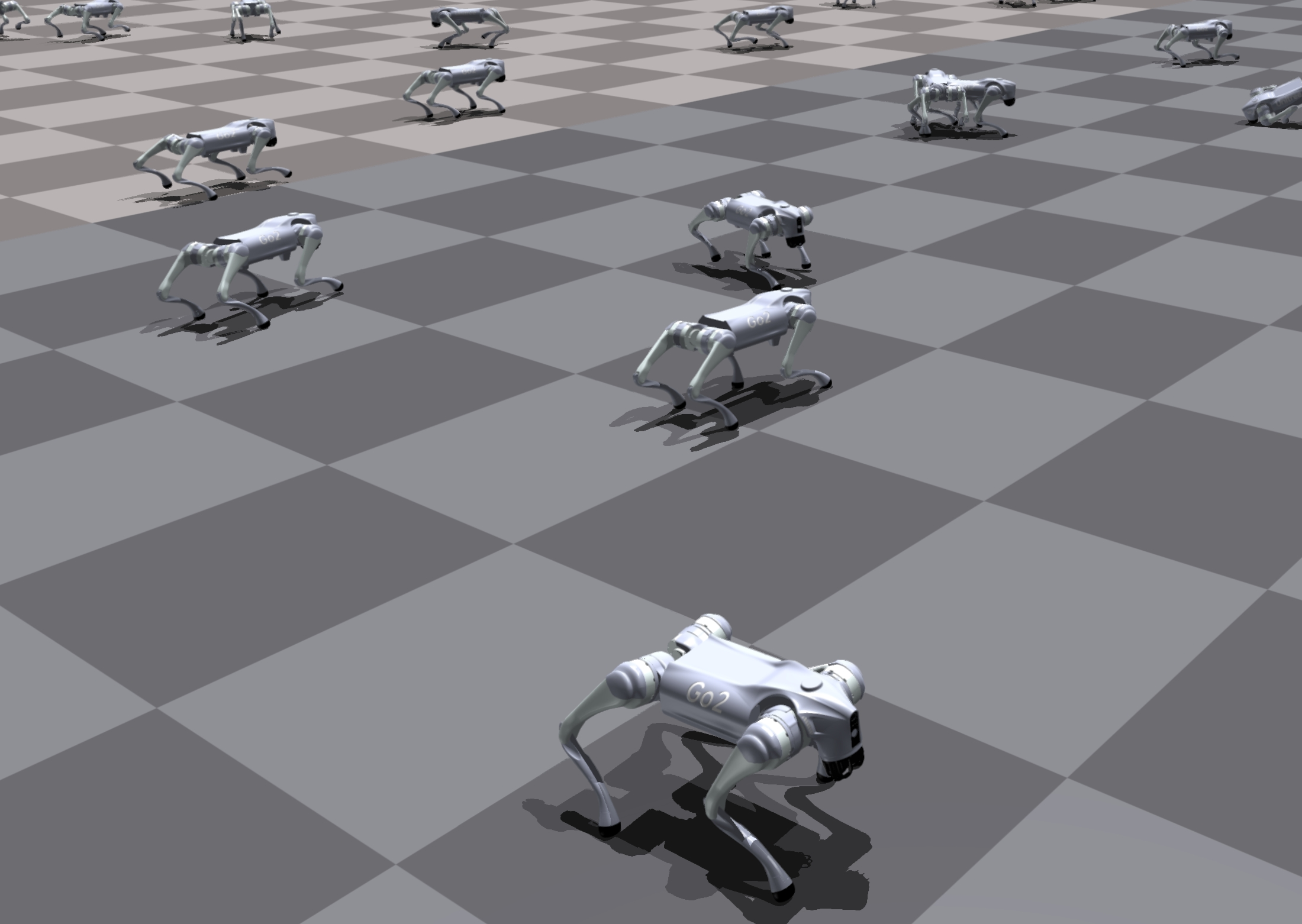}
    \caption{Zero-Shot Transfer to Unitree Go2}
    \label{fig: zero_shot}
    \vspace{-0.3cm}
\end{wrapfigure}
To evaluate generalization beyond training morphologies, we directly deploy the unified policy on Unitree Go2 without any finetuning. Despite being unseen during training, the policy achieves stable forward locomotion, demonstrating the model’s morphology-agnostic generalization. Detailed results and rollout videos are available in the supplementary video. 

For locomotion task evaluation, the Mean Episode Length (MEL), Locomotion Velocity Tracking (LVT), and Actuator Variation Threshold (AVT) metrics quantify task completion quality. As Table \ref{tab: go2} indicates, the statistically insignificant differences of CR-DP+RA in MEL ($0.62\sim3.2\%$), LVT ($0.22\sim4.1\%$), and AVT ($0.23\sim5.3\%$) between configurations a1 and go2 confirm successful zero-shot policy transfer to the go2 morphology. Also, the simulation outcomes illustrated in Figure \ref{fig: zero_shot} confirm successful zero-shot transfer to the go2 configuration, evidencing strong generalization capability of our policy framework. While Table \ref{tab: go2} reveals a substantial disparity in Average Return (AR) between CR-DP+RA implementations for a1 (387.57) and go2 (295.96), our analysis identifies this discrepancy as primarily attributable to differing desired base height parameters in the base height tracking reward module. Due to morphological variations between configurations, the optimal base height settings should be configuration-specific. After normalizing for this parameter by subtracting the base height reward component, the adjusted AR values for CR-DP+RA become 306.69 (a1) and 294.28 (go2), indicating comparable performance when accounting for morphological differences. This adjustment validates that the observed AR gap stems from parameter specification rather than fundamental policy limitations.

\begin{table}[H]
    \vspace{-0.2cm}
    \centering
    \begin{tabular}{llcccc}
        \hline
        \multirow{2}{*}{Group} & \multirow{2}{*}{Method} & \multicolumn{4}{c}{Metrics} \\
        \cline{3-6}
         & & \textbf{AR↑} & \textbf{MEL↑} & \textbf{LVT↑} & \textbf{AVT↑} \\ \hline
        \rowcolor{yellow!30}
        \multirow{2}{*}{a1} & CR-DP+RA & 387.57$\pm$6.15 & 19.60$\pm$0.19 & 5.42$\pm$0.41 & 4.32$\pm$0.41 \\ \cline{2-6}
         & CR-DP & 325.24$\pm$8.63 & 19.14$\pm$0.36 & 5.22$\pm$0.72 & 4.14$\pm$0.58 \\ \hline
        \rowcolor{yellow!30}
        \multirow{2}{*}{go2} & CR-DP+RA & 295.96$\pm$5.81 & 18.98$\pm$0.33 & 5.20$\pm$0.74 & 4.09$\pm$0.60 \\ \cline{2-6}
         & CR-DP & 256.05$\pm$5.72 & 18.40$\pm$0.37 & 5.02$\pm$0.91 & 3.93$\pm$0.74 \\ \hline
    \end{tabular}
\vspace{0.2cm}
\caption{Performance Comparison of quadruped a1 and go2}
\label{tab: go2}
\vspace{-0.2cm}
\end{table}

\section{RL residual adaptation Detail}

While diffusion models provide strong priors for action generation, they may lack task-awareness or fail to account for nuanced terrain interactions. To complement the generative prior, we introduce a residual policy trained via reinforcement learning, detailed below.

Our implementation leverages NVIDIA IsaacGym's GPU-accelerated parallel simulation environment specifically designed for robotic learning. The system architecture integrates with the established from \texttt{humanoid\_gym}~\citep{gu2024advancing} and \texttt{legged\_gym}~\citep{rudin2022learning} open-source frameworks, implementing a scalable reinforcement learning pipeline based on Proximal Policy Optimization (PPO).

\subsection{Environment Setup}
The policy observations $\bm{o}$ for the robot include velocity control commands and proprioceptive sensory data with gait cycle parameters (excluding wheeled locomotion) in the past 10 steps. Specifically:
\begin{itemize}[leftmargin=12pt]
    \item Proprioceptive Data: 
    \begin{itemize}
        \item Robot pose (orientation) and angular velocity measured by the IMU.
        \item Joint angles and angular velocities of all robotic limbs.
    \end{itemize}
    \item Velocity Control Commands:
    \begin{itemize}
        \item Linear velocity commands in the XY-plane (Cartesian coordinates).
        \item Angular velocity command around the Z-axis (yaw direction).
    \end{itemize}
    \item Gait Cycle:\\
    A time-dependent periodic signal defined by parametric sinusoidal curves:
   $$
   \text{Gait}(t) = \left\{
     \begin{aligned}
       &\sin\left(\frac{2\pi t}{T}\right)\\
       &\cos\left(\frac{2\pi t}{T}\right)
     \end{aligned}
   \right.
   $$
   where $ T $ is the gait period (time to complete one cycle),$ t $ represents the current time step.
\end{itemize}
The PPO hyperparameters are detailed in Table \ref{tab: ppo_training_params}, with a notable configuration choice of disabling value prediction clipping. This design decision stems from our hybrid architecture that integrates a pretrained diffusion model as prior guidance. The pre-trained dynamics awareness enables more stable value function initialization compared with conventional scratch training paradigms.

\begin{table}[htbp!]
    \centering
    \begin{tabular}{@{}l|l@{}}
        \toprule
        \textbf{PPO Parameter} & \textbf{Value} \\ \midrule
        Desired KL & 0.01 \\
        Learning Rate & 4e-4 \\
        Discount Factor & 0.99 \\
        Lambda(GAE) & 0.95 \\
        Mini Batches & 4 \\
        Learning Epochs & 5 \\
        Entropy Loss Scale & 0.001 \\
        PPO Clip Range & 0.2 \\
        Values Predicted Clip   & False \\
        Residual Coeff & 0.2 \\
        Max Iterations & 10001 \\
        Rollouts & 24 \\
        \bottomrule
    \end{tabular}
    \vspace{0.2cm}
    \caption{Training Parameters for PPO Unified EDM}
    \label{tab: ppo_training_params}
\end{table}
Table \ref{tab: actor_critic_model} presents our heterogeneous actor-critic architecture featuring a unified actor network parameters and four specialized critic network parameters.
\begin{table}[htbp!]
\vspace{-0.5cm}
    \centering
    \begin{tabular}{@{}l|l@{}}
        \toprule
        \textbf{Model - Critic Biped} & \\
        Activations & \texttt{["elu", "elu", "elu", "linear"]} \\
        Hidden Dims & \texttt{[512, 256, 128]} \\ \midrule
        \textbf{Model - Critic Biped Wheel} & \\
        Activations & \texttt{["elu", "elu", "elu", "linear"]} \\
        Hidden Dims & \texttt{[512, 256, 128]} \\ \midrule
        \textbf{Model - Critic Humanoid} & \\
        Activations & \texttt{["elu", "elu", "elu", "linear"]} \\
        Hidden Dims & \texttt{[512, 256, 128]} \\ \midrule
        \textbf{Model - Critic Quadruped} & \\
        Activations & \texttt{["elu", "elu", "elu", "linear"]} \\
        Hidden Dims & \texttt{[512, 256, 128]} \\ \midrule
        \textbf{Model - Actor} & \\
        Log Std Max & 4.0 \\
        Log Std Min & -20.0 \\
        Std Init & 1.0 \\
        Activations & \texttt{["elu", "elu", "elu", "linear"]} \\
        Hidden Dims & \texttt{[512, 256, 128]} \\
        \bottomrule
    \end{tabular}
    \vspace{0.2cm}
    \caption{Actor-Critic Model Parameters}
    \label{tab: actor_critic_model}
    \vspace{-0.7cm}
\end{table}

\subsection{Reward Design}
Our reward function comprises three coordinated components systematically designed for robust policy learning:
\begin{itemize}[leftmargin=12pt]
    \item Task-specific objectives governing locomotion performance, shown in Table \ref{tab: task_reward}
    \item Morphology-aware regularization terms addressing physical constraints , shown in Table \ref{tab: regulazition_reward}
    \item A diffusion-guided residual penalty enforcing dynamic feasibility through pretrained motion priors:
    \begin{equation*}
        r_d(\Delta \bm{a}_t) = \alpha\|\Delta \bm{a}_t\|_1
    \end{equation*}
    where $\alpha$ is the reward coefficient of residual penalty, $\Delta \bm{a}_t$ is the residual action.
\end{itemize}

\begin{table}[h]
\renewcommand{\arraystretch}{1.3}
\centering
\scriptsize
\begin{tabular}{cccccc}
    \hline
     Reward & Expression & Biped-Wheel & Pointed-Foot-Biped & Humanoid & Quadruped \\
    \hline
    Tracking Linear Velocity & \( \text{exp}(-\|v_b^{des}-v_b\| \times \sigma) \) & 4.0 & 2.0 & 2.0 & 6.0 \\
    Tracking Angular Velocity & \( \text{exp}(-\|\Omega_b^{des}-\Omega\| \times \sigma) \) & 2.0 & 1.5 & 1.5 & 5.0 \\
    Base Height & \( \text{exp}(-\|h_b^{des}-h_b\| \times 100)\) & - & 2.0 & 4.0 & 6.0 \\
    Orientation & \( \text{exp}(-\|\theta_b^{des}-\theta_b\| \times 10) \) & 5.0 & 5.0 & -10.0 & 4.0 \\
    \hline
\end{tabular}
\vspace{0.2cm}
\caption{Locomotion Task Reward}
\label{tab: task_reward}
\end{table}

\begin{table}[htbp!]
\vspace{-0.5cm}
\renewcommand{\arraystretch}{1.3}
\centering
\scriptsize
\begin{tabular}{cccccc}
    \hline
    Reward & Expression & Biped-Wheel & Pointed-Foot-Biped & Humanoid & Quadruped \\
    \hline
    Joint Torque & $\|\bm{\tau}\|^2$ & -1.6e-4 & -8e-5 & -8e-5 & -2e-4 \\
    Power & \( |\tau \cdot \bm{\dot q}| \) & -2e-5 & -2e-5 & - & -5e-4 \\
    Joint Vel & \( \|\bm{\dot q}\|^2 \) & -5e-4 & - & - & - \\
    Joint Acc & \( \|\bm{\ddot q}\|^2 \) & -1.5e-7 & -2.5e-7 & -2.5e-7 & -2.5e-7 \\
    Linear Velocity Z & \( \|v_b^z\|^2 \) & -0.3 & -0.5 & -2.0 & -2.0 \\
    Angular Velocity XY & \( \|\Omega_{xy}\|^2 \) & -0.3 & -0.05 & -0.05 & -0.1 \\
    Action Smoothness & \( \|a_{k+1} + a_{k-1}-2a_k\|^2 \) & -0.03 & -0.01 & -0.01 & -0.02 \\
    Action Rate & \( \|a_{k+1} - a_k\|^2 \) & -0.03 & -0.01 & -0.01 & -0.02 \\
    Collision & \( \|\bm{F}_c\| > 0. \) & -0.1 & -0.02 & -1.0 & -10.0 \\
    Contact Force & $\text{clip}\{\|\bF_{l}\|_2 + \|\bF_{r}\|_2 - F_{max}\}_{0}^{400}$ & -0.1 & -0.1 & - \\
    Default Joint Position & \( \|\bm{q}-\bm{q}_0\| \) & -0.05 & - & 3.0 & 2.0 \\
    Foot Distance & \( \text{clip}(\bm{d}_{foot}^{min} - \bm{d}_{foot}) \) & -100 & -100 & -100 & - \\
        
    Nominal Foot Height & \( \text{exp}(-\|h_{foot}^{des}-h_{foot}\|^2 \times 200) \) & 4.0 & - & - & 3.0 \\
    \hline
\end{tabular}
\vspace{0.2cm}
\caption{Regularization Reward}
\label{tab: regulazition_reward}
\end{table}

\end{document}